\documentclass{article} 
\usepackage{iclr2024_conference,times}

\usepackage{amsmath,amsfonts,bm}

\def\eqref#1{equation~\ref{#1}}

\def\1{\bm{1}}

\DeclareMathAlphabet{\mathsfit}{\encodingdefault}{\sfdefault}{m}{sl}
\SetMathAlphabet{\mathsfit}{bold}{\encodingdefault}{\sfdefault}{bx}{n}

\usepackage[
hidelinks,
colorlinks=true,
citecolor=green,
urlcolor=magenta
]{hyperref}
\usepackage{url}
\usepackage{graphicx}
\usepackage{tikz}
\usetikzlibrary{shapes, arrows, positioning, calc}
\usepackage[linesnumbered,ruled,vlined]{algorithm2e}
\usepackage{svg}
\usepackage{subfig}
\graphicspath{{./img/}}
\newtheorem{theorem}{Theorem}

\title{On the Necessity of Metalearning: \\Learning Suitable Parameterizations for Learning Processes}

\iclrfinalcopy

\author{Massinissa Hamidi \\
Laboratoire IBISC\\
Université Évry Paris-Saclay\\
\texttt{massinissa.hamidi@univ-evry.fr} \\
\AND
Aomar Osmani \\
Laboratoire LIPN-UMR CNRS 7030 \\
Université Sorbonne Paris Nord \\
\texttt{aomar.osmani@lipn.univ-paris13.fr} \\
}

\begin{document}

\maketitle

\begin{abstract}
In this paper we will discuss metalearning and how we can go beyond the current classical learning paradigm.
We will first address the importance of inductive biases in the learning process and what is at stake: the quantities of data necessary to learn.
We will subsequently see the importance of choosing suitable parameterizations to end up with well-defined learning processes. Especially since in the context of real-world applications, we face numerous biases due, e.g., to the specificities of sensors, the heterogeneity of data sources, the multiplicity of points of view, etc.
This will lead us to the idea of exploiting the structuring of the concepts to be learned in order to organize the learning process that we published previously.
We conclude by discussing the perspectives around parameter-tying schemes and the emergence of universal aspects in the models thus learned.
\end{abstract}

\section{Introduction}
Metalearning (learning-to-learn) offers promising levels of flexibility and generalization while reducing the quantities of data needed to learn (or adapt). Few-shot and zero-shot learning are examples of metalearning approaches that allow easy adaptation to new tasks (or domains), using few examples for the former or no examples at all for the latter.

Metalearning involves the study of regularities (structural dependencies) across models and tasks, where “\textit{task}” is taken in its broader sense and includes
\begin{itemize}
    \item the classical learning tasks, e.g., image classification and segmentation, activity recognition from on-body sensor deployments, etc.;
    \item robot configurations, e.g.,~\cite{cully2015robots};
    \item topologies of sensor deployment, e.g.,~\cite{hamidi2021human};
    \item multiple views (or perspectives) on a given phenomena, e.g.,~\cite{hamidi2020multi};
    \item clients in a federated deployment, e.g.,~\cite{hamidi2022context};
\end{itemize}

What characterizes a task is the tailored family of inductive biases (search or representation) that makes the learning process converge into a satisfactory solution.
The study of regularities revolves around reasoning about families of inductive biases that allow finding suitable parameterizations for the learning process.

\subsection{Inductive Biases are a Critical Pillar in the Learning Process}
Simply framed, inductive bias is some prior knowledge incorporated into the learning process that favors one particular solution over another.

Inductive biases are encoded directly into the architecture of the models via parameter-sharing schemes.

\paragraph{Parameter-sharing (or tying) schemes}

\begin{figure}[h!]
    \centering
    \includegraphics[width=0.7\linewidth]{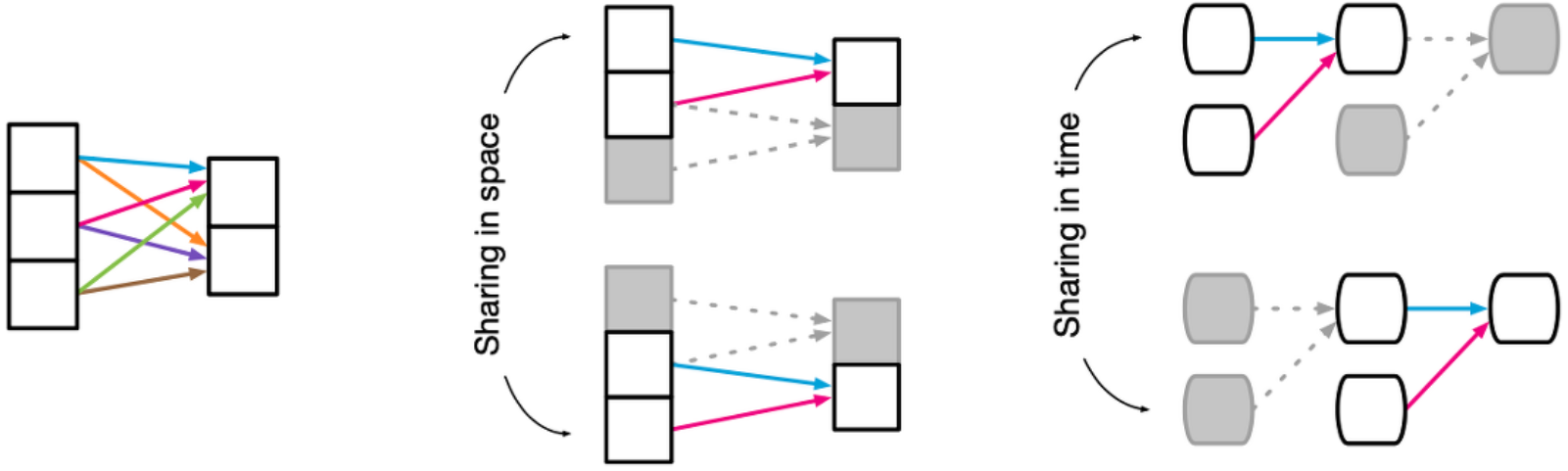}
    \caption{
        Parameter-sharing schemes.
        (a) In the fully connected layer, no sharing constraint is imposed on the weights.
        (b) In the convolutional layer, sharing is performed in the spatial dimension, where arrows with the same color indicate shared weights.
        (c) In the recurrent layer, sharing is performed across the temporal dimension.
        Figure from~\cite{battaglia2018relational}.
    }
    \label{fig:different-types-of-inductive-biases}
\end{figure}

For example, convolutional neural networks are tailored to computer vision tasks, like image segmentation and object detection, because they implement a parameter-sharing scheme in space that makes them invariant to the position in the image space.
Another example is recurrent neural networks, which are suitable for processing data that exhibit dependencies in the temporal domain. See Figure~\ref{fig:different-types-of-inductive-biases} for an illustration of these parameter-sharing schemes.

\subsection{What is at Stake? Reduced Quantities of Data and Improved Convergence Rates}
Depending on the inductive biases incorporated into the models, the resulting search space may require fewer examples to converge toward solutions.

\paragraph{Strong inductive bias}
The inductive bias is fixed once and for all.
A strong inductive bias leads to strong convergence.

Figure~\ref{fig:bias-learning-in-the-optimization-landscape} illustrates the optimization landscape of a learning problem. Highlighted in Yellow is the region we end up with when choosing models with strong inductive biases. The paths that the learning process can follow to reach a solution (depicted as a star) are constrained.

Provided that the chosen inductive bias is appropriate for the problem at hand, the learner converges to a satisfactory solution.

\paragraph{Weak inductive bias}
A model with weak inductive biases is permissive regarding the solutions that can be considered satisfactory for a task at hand.

In Figure~\ref{fig:bias-learning-in-the-optimization-landscape}, the region where the learner has to search for a solution with a model with weak inductive biases is highlighted in green. This region is much larger than the yellow region.

\subsection{Bias Learning (or Learning-to-Learn)}
\begin{figure}[h!]
    \centering
    \includegraphics[width=0.5\linewidth]{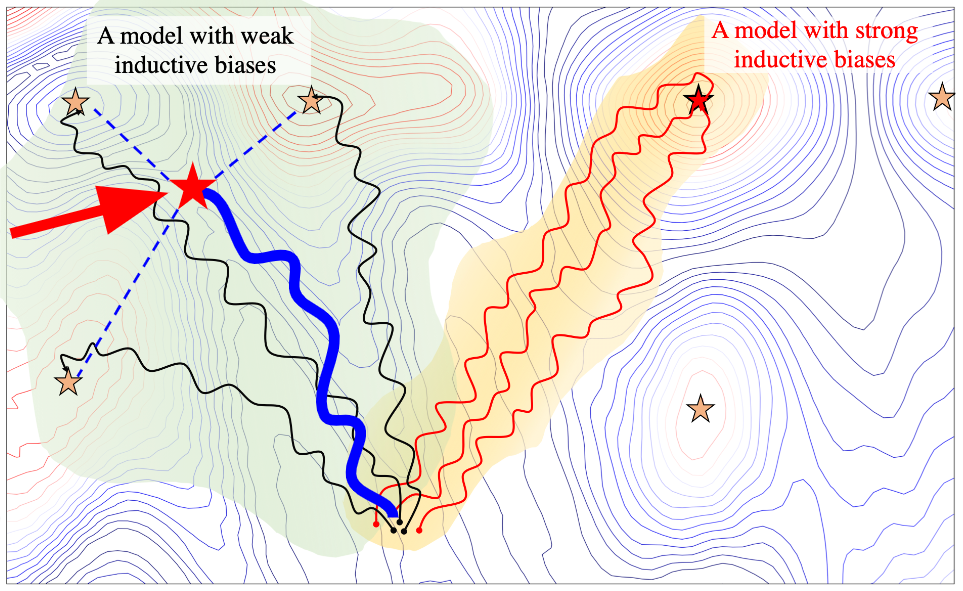}
    \caption{
        Illustration of an optimization landscape. The regions of admissible solutions induced by models with strong and weak inductive biases are highlighted in yellow and green, respectively.
        Bias learning is depicted in solid and dotted blue lines.
        Local minimizers (solutions) are depicted with stars.
        Figure adapted from~\cite{abnar2020transferring}.
    }
    \label{fig:bias-learning-in-the-optimization-landscape}
\end{figure}

Figure~\ref{fig:bias-learning-in-the-optimization-landscape} illustrates bias learning (or learning-to-learn).
Bias learning in the search space: the traditional learning process is divided into two distinct parts (1) bias learning (in bold blue) that seeks to find a particular configuration in the search space (depicted as a red star), for example, a meta-initialization in the case of gradient-based metalearning approaches~\cite{}, or a neural architecture in the case of neural architecture search approaches~\cite{}; (2) weight adaptation that seeks to find a good solution rapidly for the actual tasks at hand. Note that the stars (both red and yellow) correspond to the solutions of different tasks.

\begin{figure}[h!]
    \centering
    \includegraphics[width=0.6\linewidth]{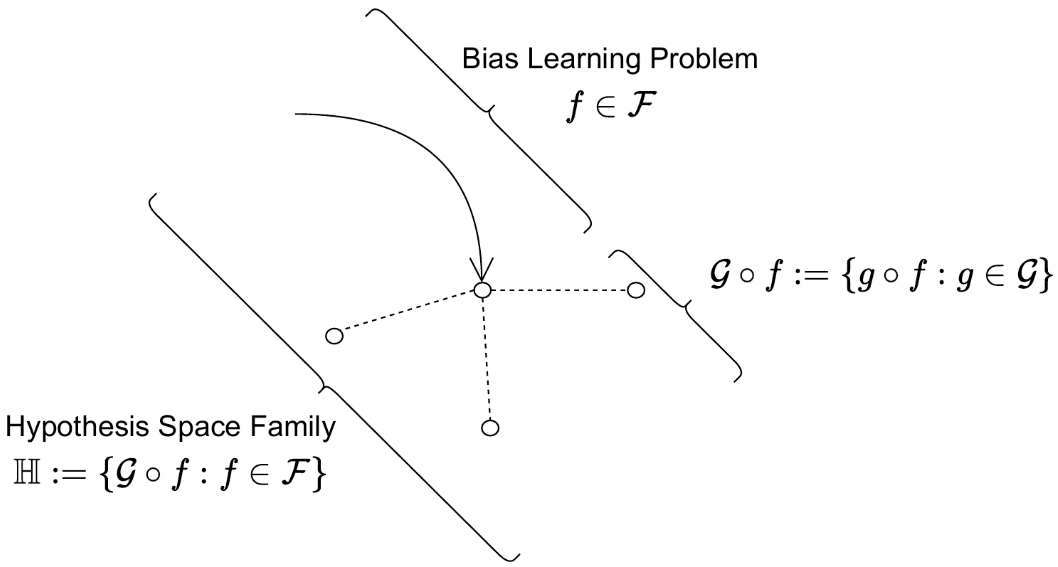}
    \caption{
        \textit{Bias learning problem}: The learning process is optimized to find a universal representation using the error signal obtained from multiple related tasks (solid line) as a first step.
        \textit{Task-specific adaptation}: The dotted lines correspond to adapting (or fine-tuning) the learned universal representation to suit specific tasks.
        Adapted from~\cite{hamidi2022metalearning} and~\cite{finn2017model}.
    }
    \label{fig:gbml-as-bias-learning}
\end{figure}

\paragraph{Example: Gradient-based metalearning}
This is a bi-level learning process where the high-level process seeks to learn a set of weights (i.e., a configuration of the neural network's weights) that can be adapted rapidly to multiple tasks.
The learning algorithm makes the model biased towards rapid adaptation by trading between the losses incurred by each individual task.
One of the prominent models of this family is Model Agnostic Metalearning (MAML)~\cite{finn2017model}.
Figure~\ref{fig:gbml-as-bias-learning} illustrates a schematic representation of gradient-based metalearning approaches.

\begin{figure}[h!]
    \centering
    \includegraphics[width=0.5\linewidth]{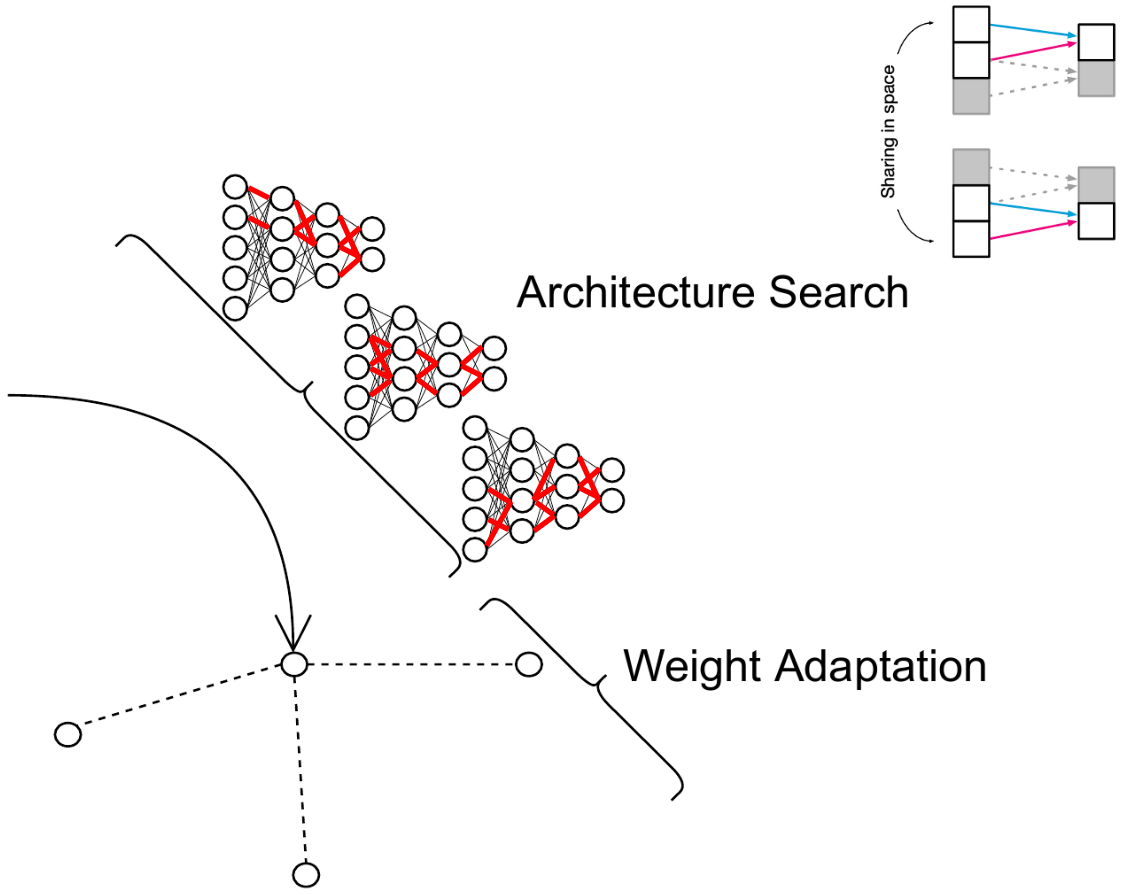}
    \caption{
        \textit{Architecture search}: in the case of neural networks, this step corresponds to finding an appropriate inductive bias. For example, a weight-tying scheme (depicted in red) that performs the convolution operation.
        At the end of this step, we get a description of the network structure but not the actual values that are assigned to the weights.
        \textit{Weight adaptation}: this step corresponds to taking the learned architecture, i.e., the network structure, and learning the actual values of the neurons that will ultimately perform the task at hand.
        Figure adapted from~\cite{hamidi2022metalearning}
    }
    \label{fig:neural-architecture-search-as-a-bias-learning}
\end{figure}

\paragraph{Example: Neural architecture search}
It corresponds to a bi-level learning process where the high-level process tries to find an appropriate architecture for the learning model. The low-level process takes the architecture and adapts its weights to a specific task.
Figure~\ref{fig:neural-architecture-search-as-a-bias-learning} shows a schematic representation of neural architecture search where the bi-level process of bias learning is highlighted.

From the perspective of parameter-tying schemes, neural architecture search looks for an optimal parameter-tying scheme that encodes a form of inductive bias.
The high-level process corresponds to finding the right architecture that encodes an appropriate inductive bias, while the low-level process learns the actual values assigned to the architecture parameters.
One key difference is that the architecture search step finds how parameters are tied together rather than the specific values assigned to these parameters.

Note the existence of training-free neural architecture search approaches, i.e., no adaptation is needed using the actual learning examples as the architectures (specific weight-tying schemes) are tailored to the specific task with random values for the weights.
For example, \cite{lin2021zen} and \cite{gaier2019weight}.
Neural architecture search was leveraged in~\cite{hamidi2020data} for human activity recognition, where different configurations of an on-body sensor deployment are dealt with using metalearned neural architectures.

\section{On the Importance of Choosing Suitable Parameterizations for Well-Conditioned Learning Processes}
Various biases~\footnote{Not to be confused, here, with inductive biases.} arise in the context of real-world applications like Internet of Things applications (§\ref{sec:biases-in-real-world-applications}). A list of biases is studied in the case of human activity recognition in~\cite{hamidi2021human}.
Naive machine learning parameterizations that ignore these biases lead to ill-conditioned learning problems (§\ref{sec:ignoring-biases-leads-to-ill-conditioned-learning-problems}).
It is, therefore, essential to appropriately choose suitable parameterizations that consider these biases leading to well-conditioned learning problems (§\ref{sec:appropriately-choosing-suitable-parameterizations}).

\subsection{Biases Arise in the Context of Real-World Applications}
\label{sec:biases-in-real-world-applications}
For example, let's consider human activity recognition from a set of on-body sensors capturing multiple modalities and located in various body positions. Usually, the datasets featuring such settings contain data (mostly time series) collected from users during multiple data collection sessions (spanning many weeks and even months).

Various problems arise in these contexts, leading to the fundamental question ``\textit{How to learn in such contexts?}'', with:
\begin{itemize}
\itemsep0.5mm
    \item Data split across the distributed data sources
    \item Temporal sequences (time series)
    \item Physical constraints on sensing and transmissions
    \item Heterogeneity of data sources
    \item Views (or perspectives) can be redundant, complementary, or, in appearance, contradictory
    \item Dynamicity of the phenomenon and sensor deployments
\end{itemize}

\subsection{Ignoring Biases Leads to Ill-Conditioned Learning Problems}
\label{sec:ignoring-biases-leads-to-ill-conditioned-learning-problems}
A problem is well-posed if its solution: exists, is unique, and depends continuously on the data~\cite{poggio1990regularization}.
However, naive machine learning parameterizations that ignore the biases arising in real-world applications lead to ill-conditioned learning problems.

\begin{figure}[h!]
    \centering
    \includegraphics[width=0.5\linewidth]{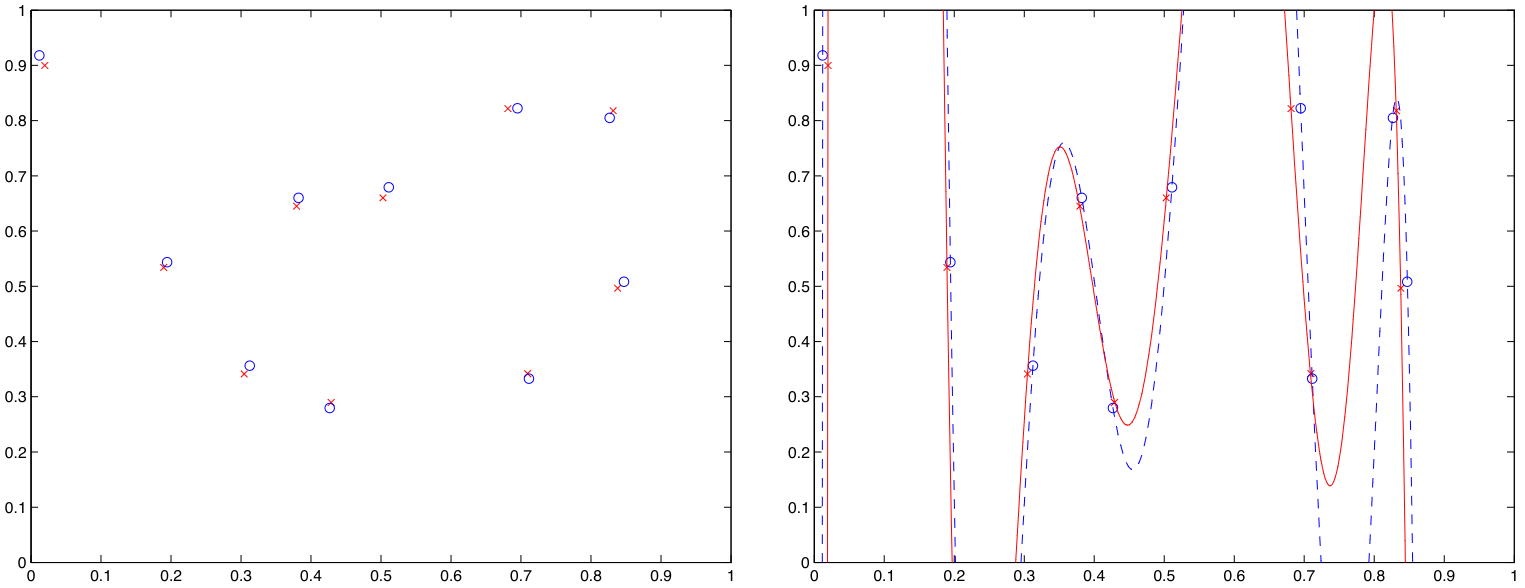}
    \includegraphics[width=0.34\linewidth]{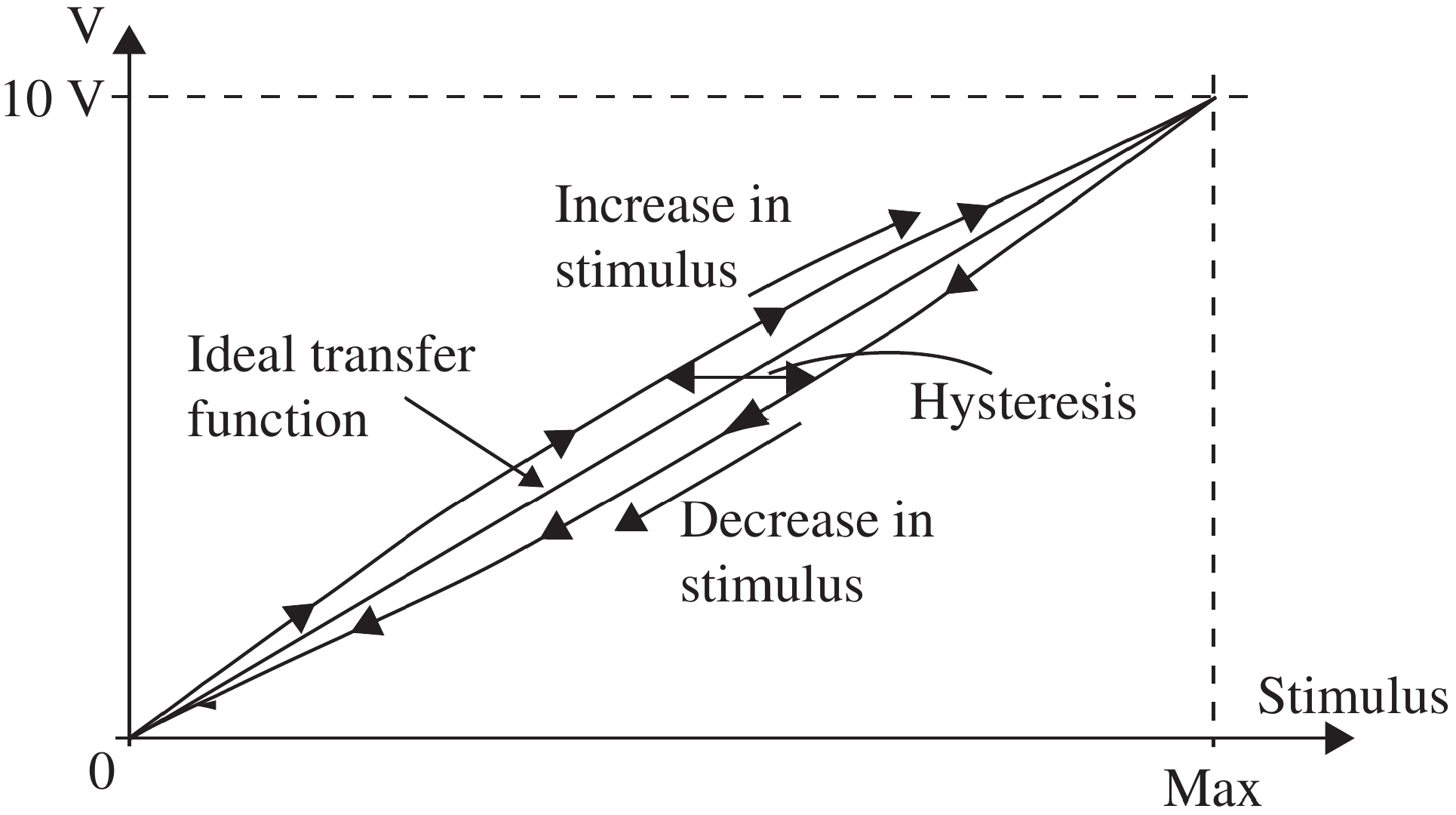}
    \caption{
        (a) set of 10 training points (blue circles \textcolor{blue}{$\circ$}) and their slightly perturbed counterparts due, for example, to the hysteresis phenomenon (red crosses \textcolor{red}{$\times$}).
        (b) Smoothest interpolating polynomial fit with a degree of 9.
        (c) Hysteresis in a sensor.
        The infinitesimal perturbations brought to the initial set of training points can be related to the hysteresis phenomenon during the sensing process.
    }
    \label{fig:polynomial-fit}
\end{figure}

\paragraph{Sensor specificities}
Data points are the end result of a sensing process.
This process is tainted with various biases due to the sensors' intrinsic characteristics (or specificities), including precision, repeatability, hysteresis, etc.
Figure~\ref{fig:polynomial-fit} shows the hysteresis phenomenon during the sensing process and the impact of infinitesimal variations (due, for example, to the hysteresis) on the learned theories. 

\paragraph{Sensor point-of-view is biased by its location relative to the phenomena of interest}
Infinitesimal variations in the sensor's position w.r.t. the phenomena (or physical quantity) of interest lead to different outputs of the sensing process. Ultimately, the theories learned from the outputs of these sensing processes do not reflect the true underlying phenomena of interest.
These variations are often referred to as ``\textit{model variance}'', where models with ill-conditioned hypothesis spaces tend to have high variance, i.e., small changes in the training data can lead to significantly different models. This makes the model sensitive to noise and fluctuations in the data.

\begin{figure}[h!]
    \centering
    \includegraphics[width=0.5\linewidth]{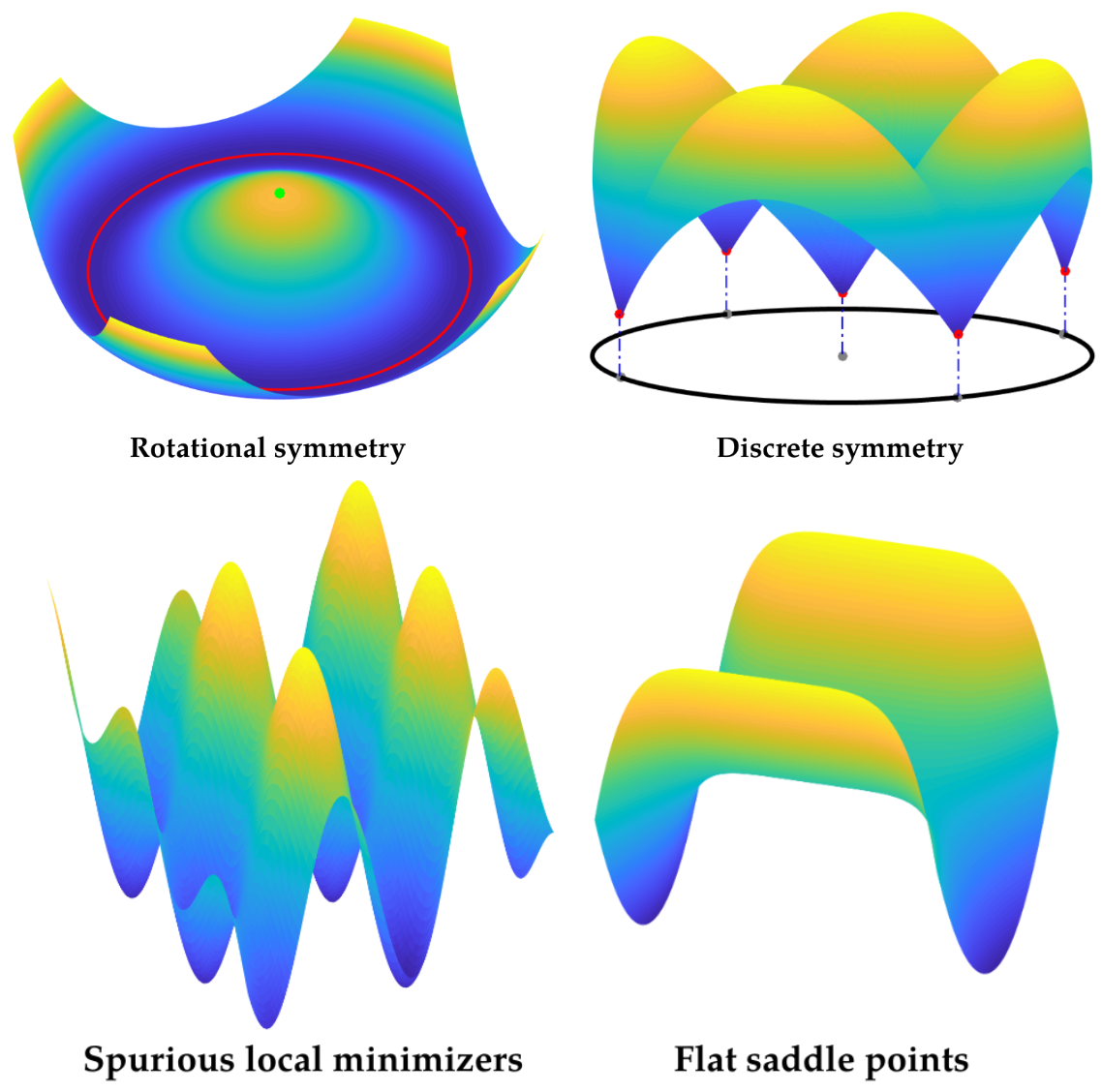}
    \caption{
        Various forms of the optimization landscape, where “symmetries of the observation models become symmetries of the optimization problem”, i.e., factors of the real world shape the optimization landscape in particular ways.
        On the one hand, \textit{Rotational symmetry} and \textit{Discrete symmetry} are examples of nonconvex problems that can be solved globally with efficient optimization methods under certain hypotheses.
        On the other hand, optimization landscapes may be endowed with \textit{Spurious local minimizers} and \textit{Flat saddle points}, which make local methods trapped near local minimizers or can stagnate near flat saddle points.
        Figure from~\cite{zhang2020symmetry}.
    }
    \label{fig:from-symmetry-to-geometry}
\end{figure}

\paragraph{Sensors point-of-views are relative to each other}
Each sensor's perspective (or view) is relative to other sensors' perspectives, meaning that they can be overlapping, contradicting, or redundant.
In this context, we end up with many distinct situations:
One of these is related to the form of the optimization landscapes, which can be impacted by various factors, e.g., symmetrical perspectives can lead to optimization challenges.
Figure~\ref{fig:from-symmetry-to-geometry} illustrates how symmetries of the real world are translated into artifacts in the optimization landscape.

Finding a good solution (or hypothesis) to the learning problem requires the traversal of the optimization landscape, which could be very difficult~\footnote{The optimization landscape is the geometry of the network’s loss function or the response of the network’s loss function when its weight values are adjusted.}.
Nevertheless, the very often non-convex optimization landscape is explored using local search heuristics, as simple as gradient descent, achieving remarkable state-of-the-art results.
The difficulty of this traversal depends on the properties of the optimization landscape~\cite{dauphin2014identifying,li2018visualizing,ahmed2019understanding}.
Indeed, the optimization landscape might be chaotic with shallower regions of convexity, where the gradients provided by the local search heuristics are likely uninformative~\cite{li2018visualizing}.
Furthermore, authors in~\cite{dauphin2014identifying} investigated the prevalence of saddle points in high-dimensional non-convex optimization problems, which may hinder learning and make the optimization procedure take a long time to escape.
The curvature of the optimization landscape can also vary rapidly, which makes choosing a step size for the optimization procedure very difficult~\cite{ahmed2019understanding}.

\paragraph{Heterogeneity bias}
Similar to the relativity of sensors' point-of-views, fusing different local models with varying and heterogeneous learning objectives leads to difficulty in optimization.
Long lines of research have been dedicated to this problem, e.g., federated learning~\cite{konevcny2016federated,osmani2022reduction,hamidi2022context,li2020federated}

\paragraph{The concepts to learn (or classes, or labels) can overlap}
Drawing a clear separation between the concepts to learn is quite difficult in real-world applications.
In the literature, the classes are often assumed to be separable. In real-world applications, this assumption becomes strong and generates inconsistencies.
The concepts to learn exhibit dependencies.
This is linked to the semantic definition of the concept itself or the fact that a concept comprises sub-concepts (atomic non-decomposable concepts).

\begin{figure}[h!]
    \centering
    \includegraphics[width=\columnwidth]{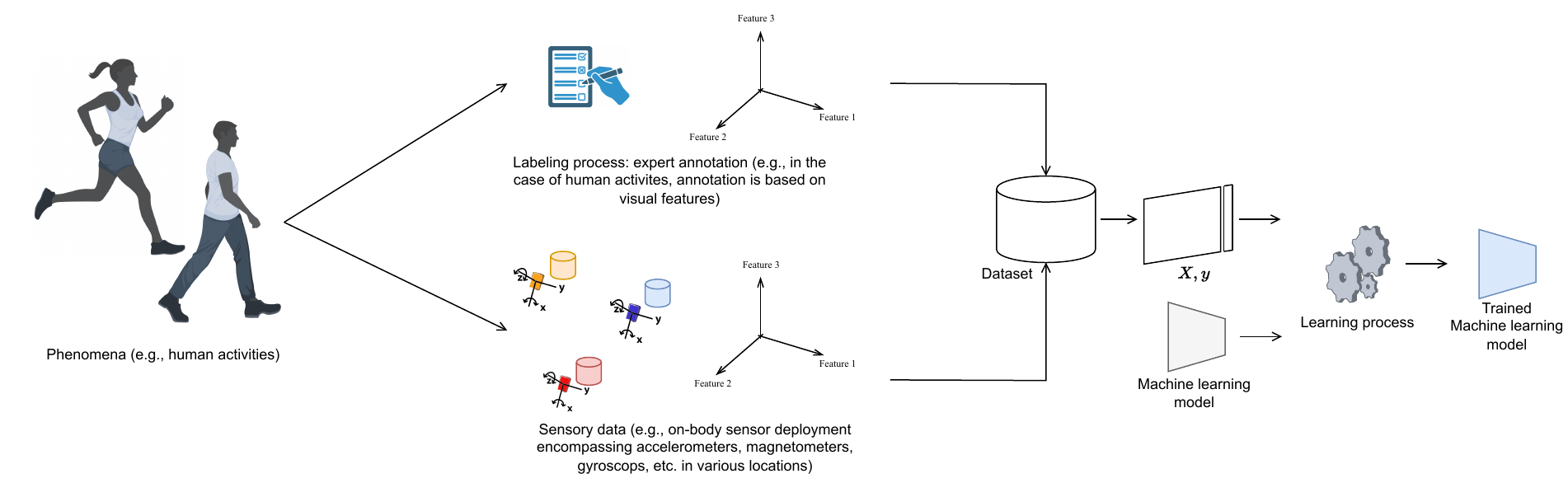}
    \caption{
        The labeling process in the machine learning process.
    }
    \label{fig:diagram-of-the-labeling-process}
\end{figure}

\paragraph{Labeling bias}
Overlap of learning examples, ambiguous definition of the labels, labels dependencies, etc.
Figure~\ref{fig:diagram-of-the-labeling-process} illustrates the labeling process within the machine learning process.
In machine learning, we monitor phenomena (e.g., human activities) using a set of modalities (e.g., accelerometers, magnetometers, etc.). We construct datasets by collecting these modalities and assigning labels to temporal segments of these modalities. Label assignment is based on domain experts who use some criteria. Then, we feed these modalities to train models that are capable of predicting the labels. This process highlights a gap between the criteria used to label the temporal segments and the modalities used to describe the phenomena.
But: How to decide which features to use for the assessment step? What happens if different types of features are used (which often happens with different expert raters)?

\paragraph{Temporal bias}
Segmentation or decomposition of the inputs.
Many types of segmentation procedures exist in the literature around activity recognition and beyond, including time-, event-, and energy-based. 
Various works studied the effects of different segment lengths on the recognition performances empirically~\cite{shoaib2016complex,banos2014window}.
Issues with time-based segmentation are not circumscribed to the choice of the segment's length but are also tightly linked to the feature extraction step.
Activities that last for variable time constitute a critical issue.
For example, fixing the segment path can result in spectral leakage that impacts the subsequent steps, which noticeably includes the feature extraction step from the spectral representation of the signal. Indeed, spectral leakage causes the spectrum to be noisy, impacting the correct determination of frequencies, etc.

\paragraph{Evaluation and neighborhood bias}
Model evaluation based on cross-validation usually relies on a random partitioning process.
The random partitioning used in the case of segmented time series introduces a neighborhood bias~\cite{hammerla2015let}. This bias consists of the high probability that adjacent and overlapping sequences (which are typically obtained with a segmentation process and that share a lot of characteristics) fall into training and validation folds at the same time.
This leads to an overestimation of the validation results and goes often disregarded in the literature.
We investigated in our previous works~\cite{osmani2017platform,osmani2017machine} the impact of such bias.

\subsection{Metalearning, or Appropriately Choosing Suitable Parameterizations}
\label{sec:appropriately-choosing-suitable-parameterizations}
if the chosen inductive bias is too strong or inappropriate for the given task, it can lead to an ill-conditioned hypothesis space.

An ill-conditioned hypothesis space is one where the model is overly complex, and the number of possible hypotheses is large relative to the amount of training data available. This can result in several issues: overfitting, high variance, computational challenges, and difficulty in optimization.

Choosing an appropriate inductive bias is crucial to strike a balance between model complexity and generalization. It involves understanding the underlying patterns in the data and selecting a hypothesis space that captures those patterns without overfitting to noise. This process often requires domain knowledge and experimentation to find the right level of model complexity for a given task.

Rather than choosing suitable parameterizations by hand, why not learn them?
In this case, metalearning corresponds to a bi-level process, where (1) the high-level process is responsible for learning an appropriate parameterization (e.g., finding an appropriate neural network’s architecture) that (2.a) eases the classical learning processes (e.g., training a neural network) at a lower level and (2.b) guides it to solutions with certain desired properties.

Indeed, without such high-level “supervision,” exploring the state space of these learning problems is deemed impractical because of their size, requiring lots of data to reach certain solutions, and suboptimal (reaching unsatisfactory solutions) because the state space is ill-conditioned.
For example, as explained in the previous section, symmetric perspectives provided by a dataset can notably lead to overly complex optimization landscapes, which are difficult to explore when searching for solutions to learning problems~\citep[see §4.6]{hamidi2022metalearning}. In this case, learning-to-learn is about finding more appropriate representations that simplify the optimization landscape and make it easier to explore. With these approaches, we can achieve dual goals: data-efficient learning processes and highly accurate and robust learning models.
For example, metalearning approaches, like gradient-based metalearning, achieve tremendous breakthroughs in various domains like computer vision (e.g., leveraging multiple heterogeneous vision tasks), especially in terms of rapid adaptation to new unseen tasks using limited quantities of data, e.g.,~\cite{nichol2018reptile,finn2017model,raghu2019rapid}.

\section{Structuring the concepts to learn}
This leads us to the study of an approach that is based on the structuring of the concepts to be learned in order to organize the learning process.

\begin{figure}[h!]
    \centering
    \includegraphics[width=0.7\linewidth]{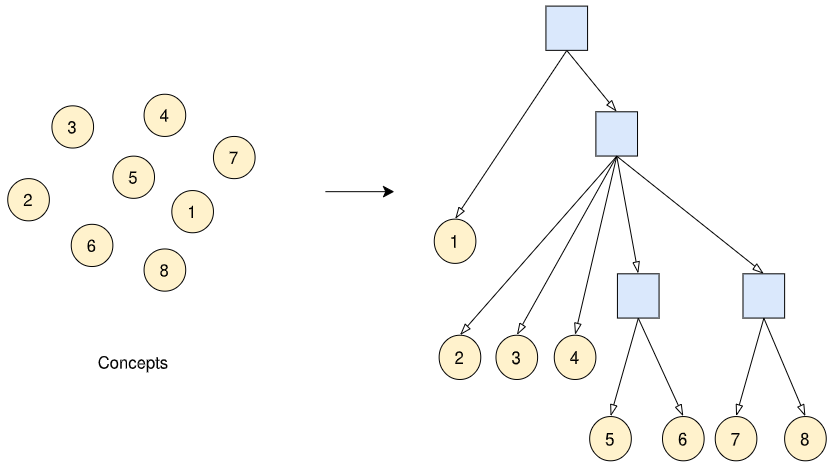}
    \caption{
        Transition from a flattened set of classes (left) to a hierarchy (right) whose structuring makes it possible to guide the learning process.
        Here, first learn to classify class 1 against the rest before learning to classify, for example, classes 5 and 6.
    }
    \label{fig:structuring-the-concepts-to-learn}
\end{figure}

This idea was proposed in~\cite{osmani2021hierarchical} and~\cite{osmani2022clustering}.
It is motivated by the dependencies (or overlaps) that exist between concepts to learn in real-world applications (§~\ref{sec:concepts-dependency-in-real-world-applications}).
Consequently, instead of treating the concepts to learn as a flattened set, their structuring (into hierarchies, for example) is exploited in order to guide the learning process (§~\ref{sec:from-flat-to-a-succession-of-learning-problems}).
Figure~\ref{fig:structuring-the-concepts-to-learn} illustrates this idea of going from a flattened set of classes to a hierarchy whose structuring makes it possible to guide the learning process.

We present in §~\ref{sec:concepts-structuring-based-on-clustering} and §~\ref{sec:concepts-structuring-based-on-transfer-affinity} two metalearning approaches whose objective is to learn to structure learning based on the structuring of the concepts to be learned.
These two approaches are based on key principles to cope with the combinatorial explosion of the high-level process of metalearning.

\subsection{Concepts dependency in real-world applications}
\label{sec:concepts-dependency-in-real-world-applications}
Drawing a clear separation between the concepts to learn is quite difficult in real-world applications.
In the literature, the classes are often assumed to be separable. In real-world applications, this assumption becomes strong and generates inconsistencies.
The concepts to learn exhibit dependencies.
This is linked to the semantic definition of the concept itself or the fact that a concept comprises sub-concepts (atomic non-decomposable concepts).

\begin{figure}[h!]
    \centering
    \includegraphics[width=0.7\linewidth]{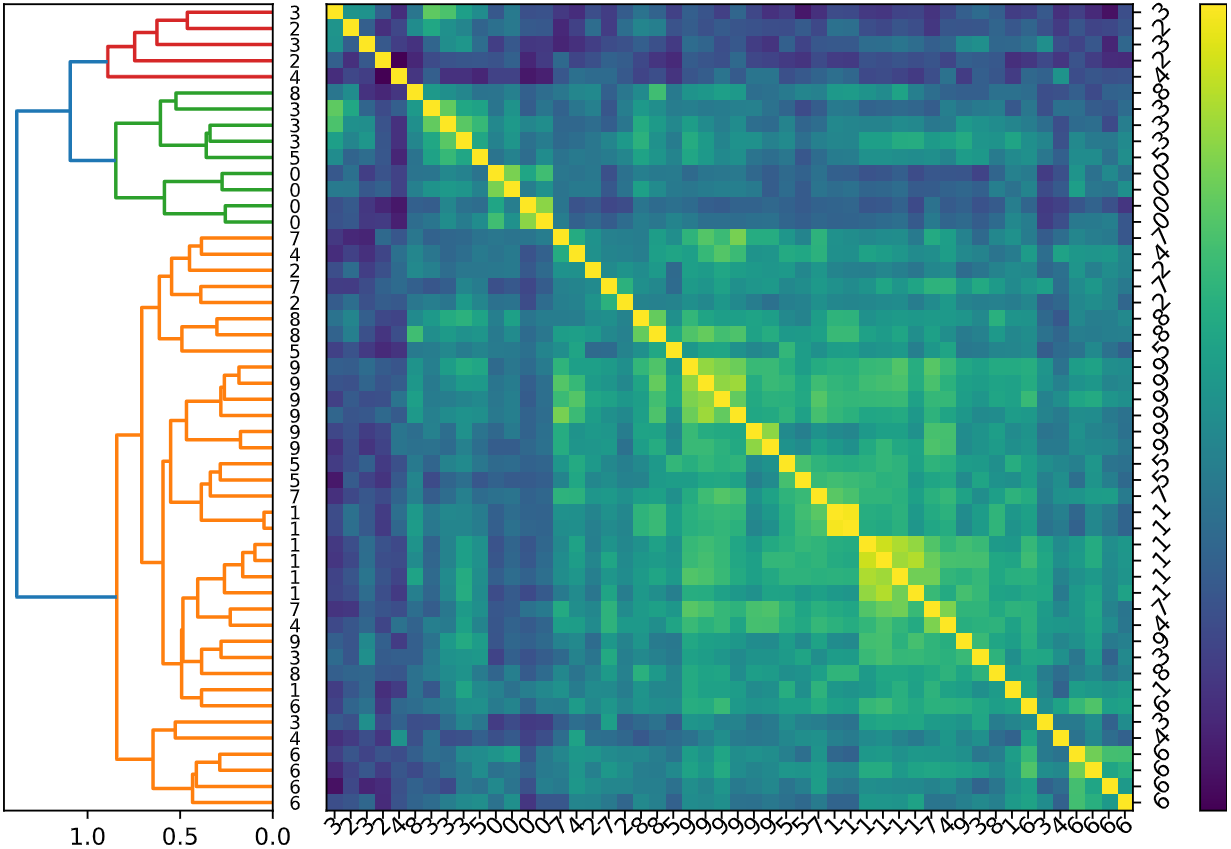}
    \caption{Dendrogram and similarity matrix computed on a subset of the MNIST dataset.}
    \label{fig:dendrogram-and-similarity-matrix-subset-of-MNIST-dataset}
\end{figure}

\paragraph{Case of the MNIST dataset}
Figure~\ref{fig:dendrogram-and-similarity-matrix-subset-of-MNIST-dataset} illustrates the various levels of dependencies among the different digits of the MNIST dataset.

\begin{figure}[h!]
    \centering
    \includegraphics[width=0.5\linewidth]{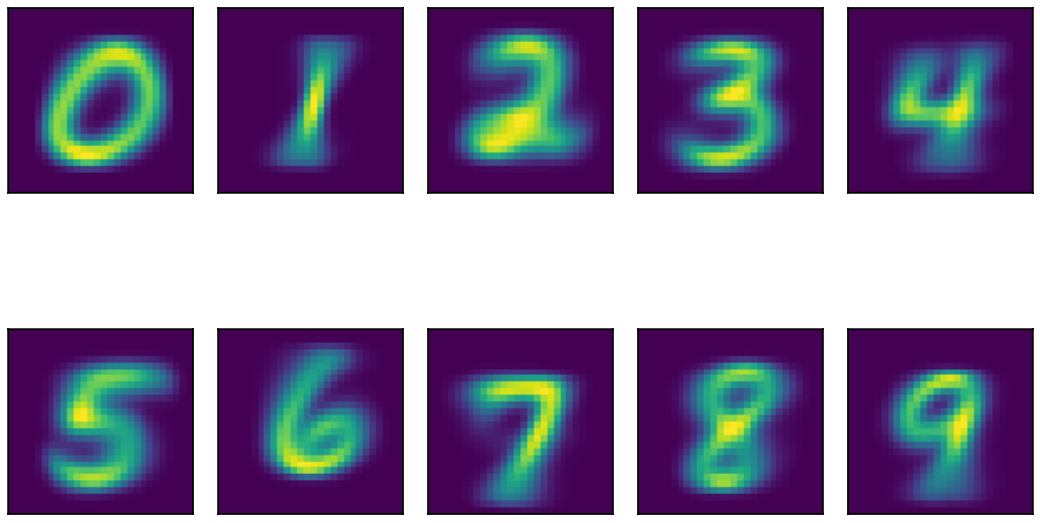}
    \includegraphics[width=0.35\linewidth]{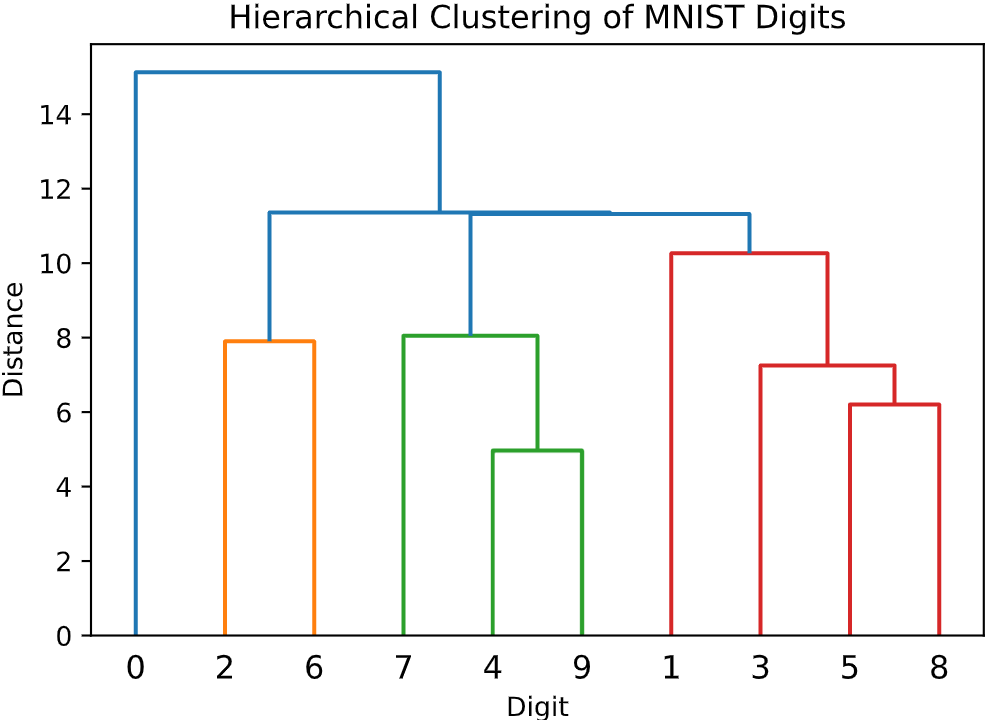}
    \caption{
        (a) Average image for each digit in the MNIST dataset.
        (b) Hierarchical clustering of the MNIST digits.
    }
    \label{fig:average-image-MNIST-digits}
\end{figure}

Figure~\ref{fig:average-image-MNIST-digits} illustrates the average image for each digit in the MNIST dataset and the corresponding structuring of the digits as obtained using a hierarchical clustering approach.

\paragraph{Case of human activity recognition}
For example, Running and Walking are seemingly two different activities (at least, they are treated as such in the vast majority of HAR applications).
But, in reality, who can draw a line between these two activities? When can we say a given data stream corresponds to Running and not Walking, and vice versa?
These two classes are, therefore, related to each other, and, as such, the models that learn these concepts potentially \textit{share a large body of aspects}.
In other words, we end up with the problem of \textit{overlapping perspectives} mentioned in the previous section.

Very often, learning the concepts one against the other is a strong simplification, which does not reflect reality. With this strong simplification that ignores the relationship (dependencies), the learning problem is made unnecessarily more challenging as we re-learn many aspects that are shared across the concepts and groups of concepts.

Instead of building models that simply consider a set of flattened concepts and ignore the dependencies across concepts and sub-concepts, we can take the other way around and leverage these dependencies to organize (or guide) the learning process according to the way these concepts are structurally related, e.g., in the form of graphs or hierarchies or even a continuous space.

This general observation shows that inductive biases needed to separate homogeneous groups of concepts recursively give better results and build hierarchical concept structure between concepts.

\subsection{From a flat learning problem to a succession of learning problems with increasing levels of difficulty}
\label{sec:from-flat-to-a-succession-of-learning-problems}
Organize the learning process into a succession of learning problems of increasing levels of difficulty according to the concepts' dependencies.

This point of view is also motivated by the natural link with learning in a student (student) where the concepts should be presented by an increasing \textit{degree of difficulty}: from the simplest concept to the most complex one. Indeed, We find that some concepts are easier to distinguish when grouped with other concepts than when each one is learned on its own. For instance, if we consider analyzing human activities through the accelerometer or heart rate, it is easier for a given learner to first separate all activities (concepts) into two main classes, e.g., activities involving large movements of the hand versus other activities, instead of separating the finer activities belonging to (or lying within) these two general classes.

\begin{figure}[h!]
    \centering
    \includegraphics[width=0.4\linewidth]{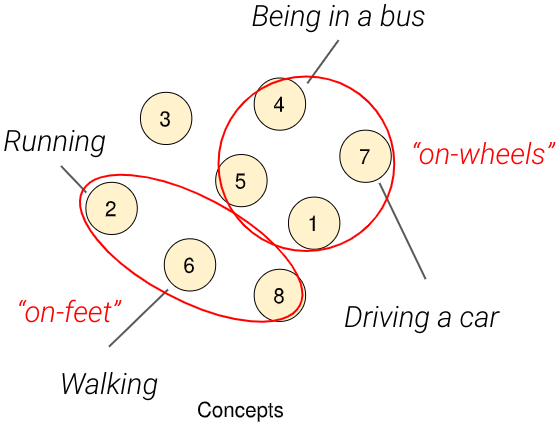}
    \caption{
        The eight (coarse-grained) concepts (or human activities) featured in the SHL dataset.
        For example, on the one hand, Running and Walking can be grouped in the ``on-feet'' activities.
        On the other hand, Being in a bus or Driving a car can be grouped in the ``on-wheel'' activities.
    }
    \label{fig:concepts-dependency-in-real-world-applications}
\end{figure}

The idea is to start with learning the biases that are appropriate to learn the groups of concepts and then leveraging what was learned to tackle the problem of the next level.
For example, it is easier to learn ``on-feet'' group of activities (such as running and walking) against an ``on-wheel'' group (like being on a bus or biking) before learning more specific concepts inside each of these groups.
See Figure~\ref{fig:concepts-dependency-in-real-world-applications}.

This way, we build a hierarchy with a succession of increasingly specialized inductive biases.

Given a set of concepts to learn, we construct an optimal hierarchy that reflects the true dependencies that exist among the concepts and finally leverage the hierarchy for an efficient learning process.

\begin{figure}[h!]
    \centering
    \includegraphics[width=0.6\linewidth]{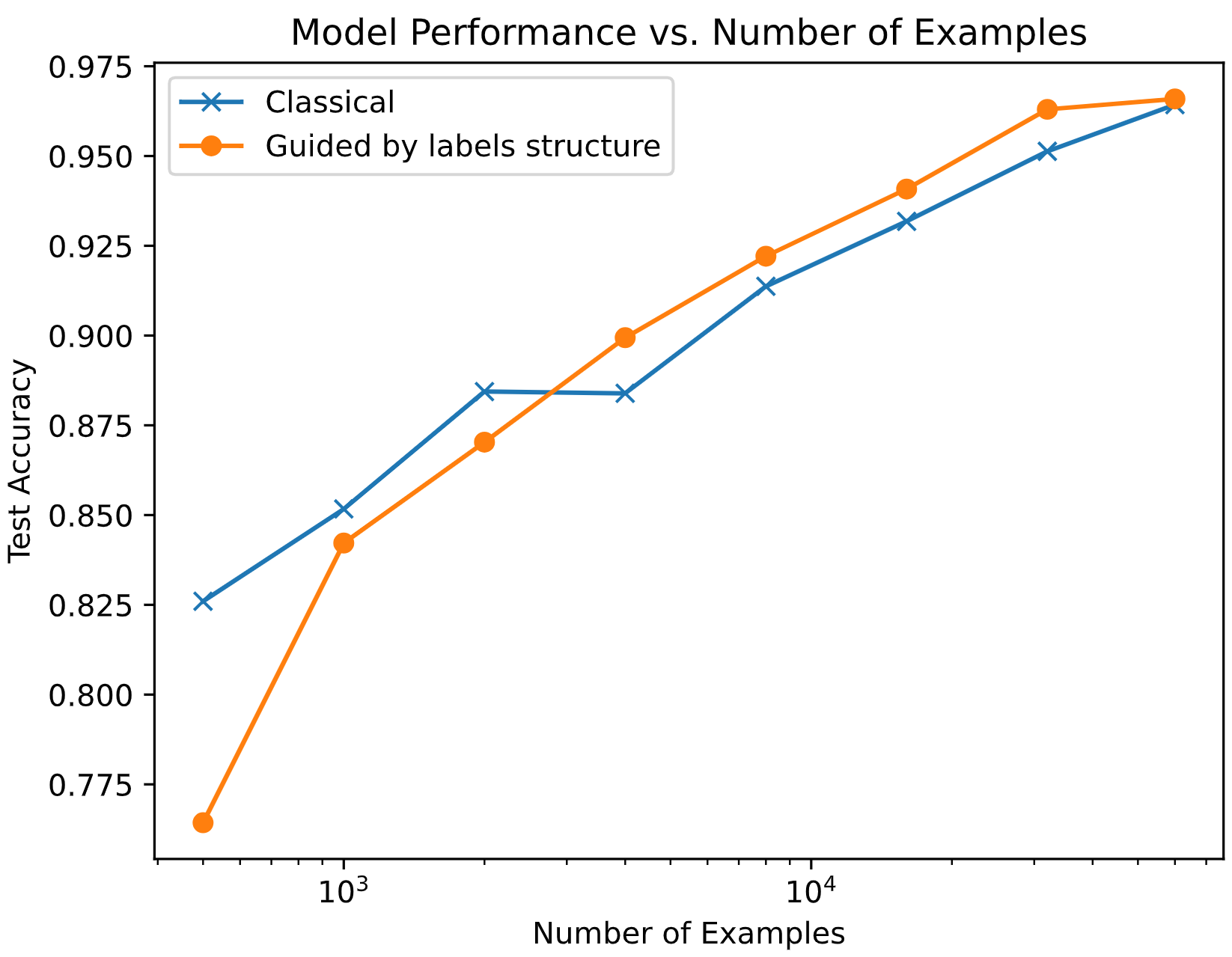}
    \caption{Model performance vs. number of examples MNIST dataset.}
    \label{fig:model-performance-vs-number-of-examples-MNIST-canonical-vs-hierarchical}
\end{figure}

\paragraph{Case of the MNIST dataset}
Here, we trained simple neural networks using, on the one hand, the classical setting and, on the other hand, the additional supervision provided by the hierarchical structure of the concepts to learn.
The hierarchical structure used to guide the learning process is the one depicted in Figure~\ref{fig:average-image-MNIST-digits}(b).
We varied the number of training examples $n$ with $n\in\{500, 1000, 2000, 4000, 8000, 16000, 32000, 60000\}$.
We use the cross-entropy loss defined as $\sum p(x)\ln q(x)$, where $p(x)$ is the probability distribution of the targets and $q(x)$ is the probability distribution of the model's predictions.
Figure~\ref{fig:model-performance-vs-number-of-examples-MNIST-canonical-vs-hierarchical} shows the model's performance as a function of the number of examples in the training set.
It compares the performance obtained using the classical learning process (all concepts to learn are flattened) against the one obtained by the learning process guided by the concepts to learn (hierarchically structured concepts).

\begin{figure}[h!]
    \centering
    \includegraphics[width=0.45\linewidth]{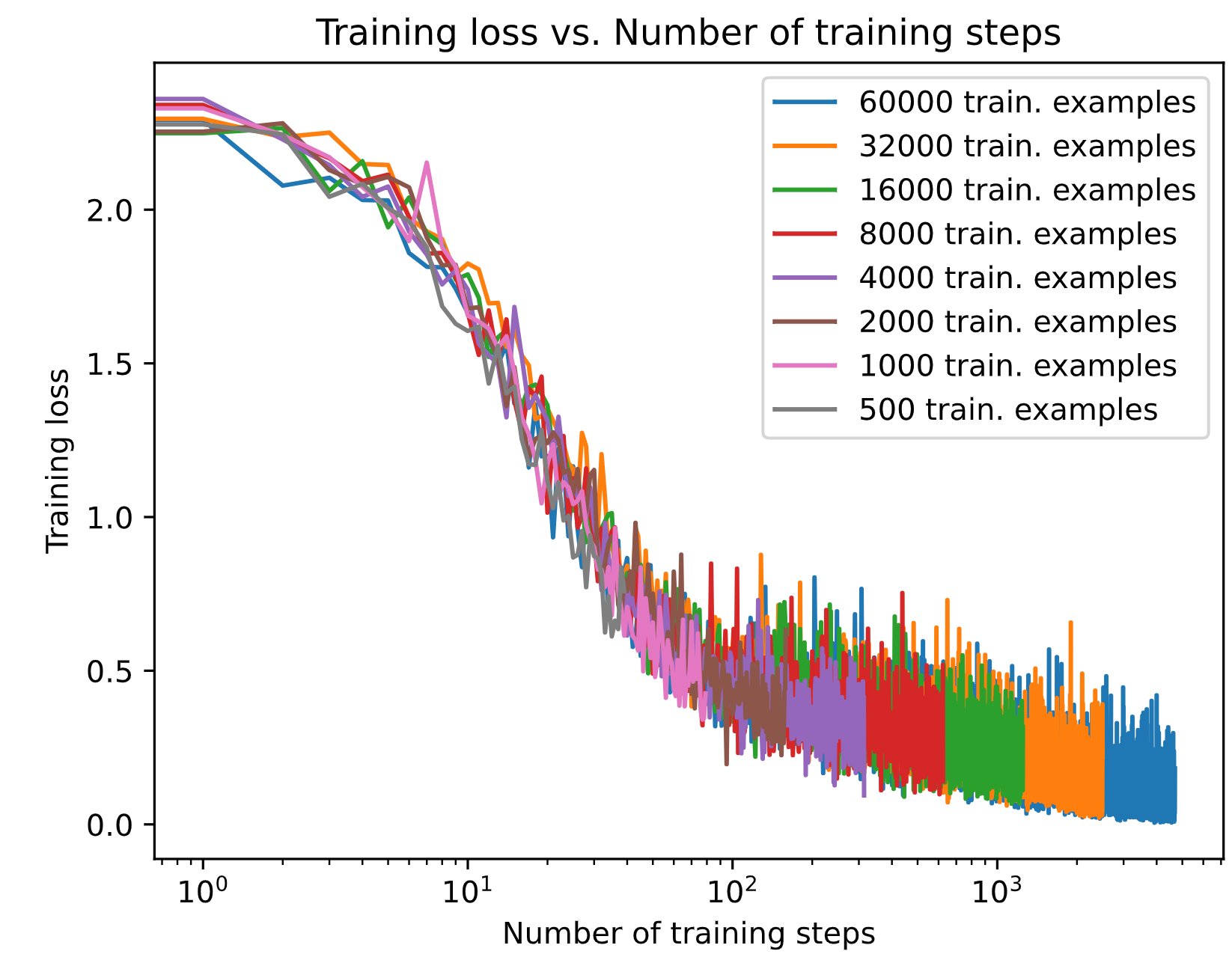}
    \includegraphics[width=0.45\linewidth]{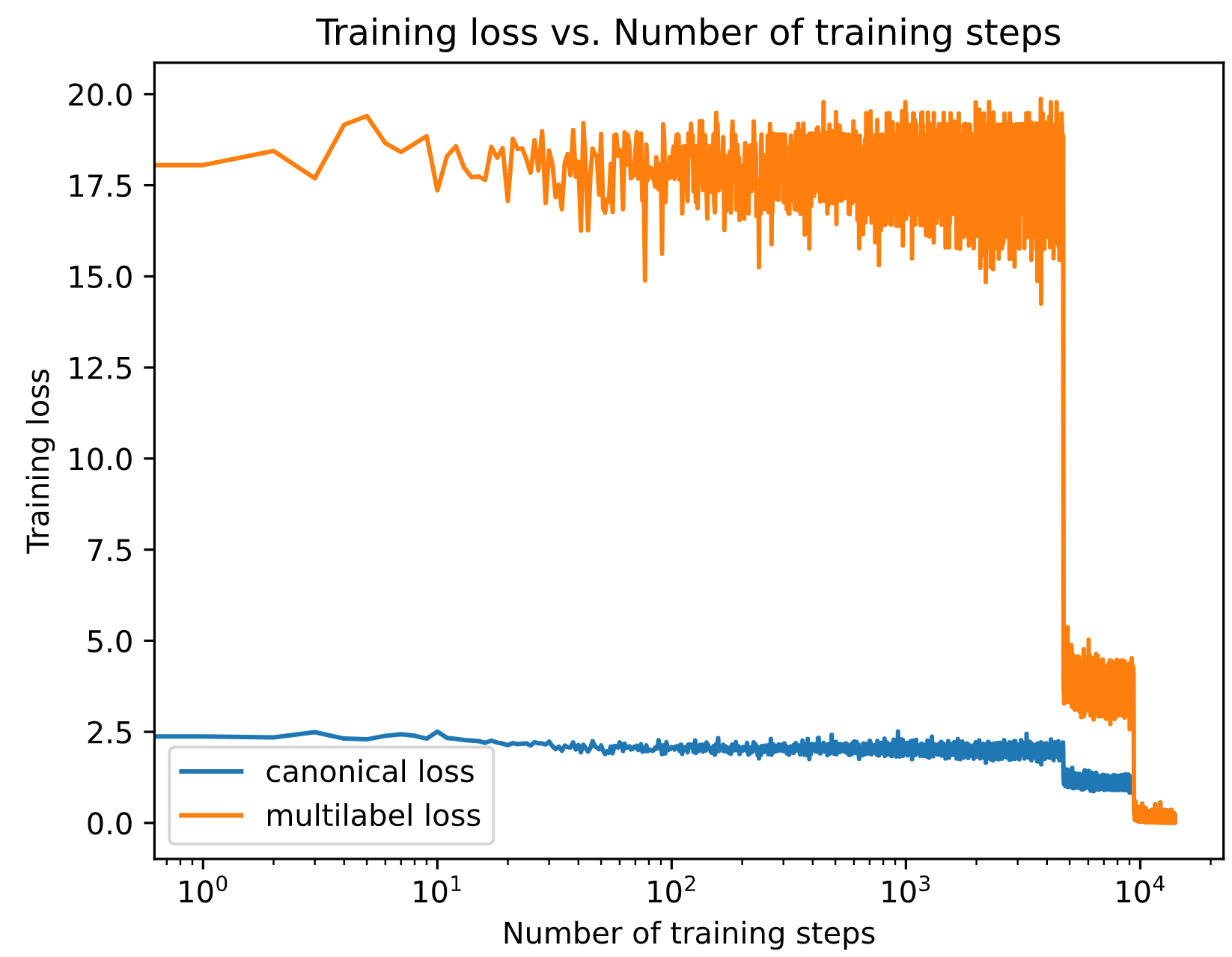}
    \caption{
        Training loss vs number of examples MNIST dataset; Number of examples in a log scale.
        (a) classical learning.
        (b) learning process guided by the structure of the concepts to learn with 60k training examples.
    }
    \label{fig:training-loss-vs-number-of-examples-canonical-and-hierarchical}
\end{figure}

Figure~\ref{fig:training-loss-vs-number-of-examples-canonical-and-hierarchical} shows the evolution of the training loss as a function of the number of training steps during the classical and hierarchically guided learning process.
In the learning process guided by the structure of the concepts to learn, we can distinguish three training phases corresponding to the three levels of the concepts' hierarchical structure depicted in Figure~\ref{fig:average-image-MNIST-digits}(b).
As a first step, we can see that the level of the multilabel loss signal saturates and remains between 15 and 20 until the end of the first level of the learning process.
It then goes down and reaches the same level as the canonical loss.
The intermediate weight configurations explored using the canonical and multilabel losses during this process are different.
The final solutions reached by these processes need further investigation: are they the same, equivalent, or linked by valleys of low loss as in~\cite{mirzadeh2020linear}? 

\subsubsection{… but the Size of the Search Space is Impractical}
Generating and testing all the configurations that the concepts to learn can take is not feasible in practice.

Indeed, a naive approach consists of building all the combinations of concepts to check for which groups of classes the quality of the learning is optimal and to start again recursively this approach until the concepts are totally separated from each other. However, this approach faces a combinatorial explosion of the number of cases that should be treated.
To better illustrate the complexity of this problem, we propose a recurrence relation involving binomial coefficients for calculating the total number of tree hierarchies for a total number of $n$ concepts.
\begin{theorem}
Let $L(n)$ be the total number of trees for the $n$ atomic concepts.
The search space size for these concepts satisfies a recurrence relation defined as:
\end{theorem}
{
$$L(n)= {n-1 \choose n-2}L(n-1)L(1)
    +2\sum_{i=0}^{n-3} {n \choose i} L(i+1) L(n-i-1) $$
}

For example, with $8$ coarse-grained concepts, the size of the search space is $L(8) = 660,032$.

The idea is, therefore, to learn appropriate parameterization for the base learning process.
In the following, we will evoke two metalearning approaches (§~\ref{sec:concepts-structuring-based-on-clustering} and §~\ref{sec:concepts-structuring-based-on-transfer-affinity}) that go into the sense of learning appropriate organization of the learning process based on the concepts to learn.

\subsection{Concepts Structuring Based on Clustering}
\label{sec:concepts-structuring-based-on-clustering}
\begin{figure}[h!]
\captionsetup[subfigure]{labelformat=empty}
\centering
\sffamily
\subfloat[]{
    \def\svgwidth{1.3\columnwidth}
    \resizebox{110mm}{!}{
       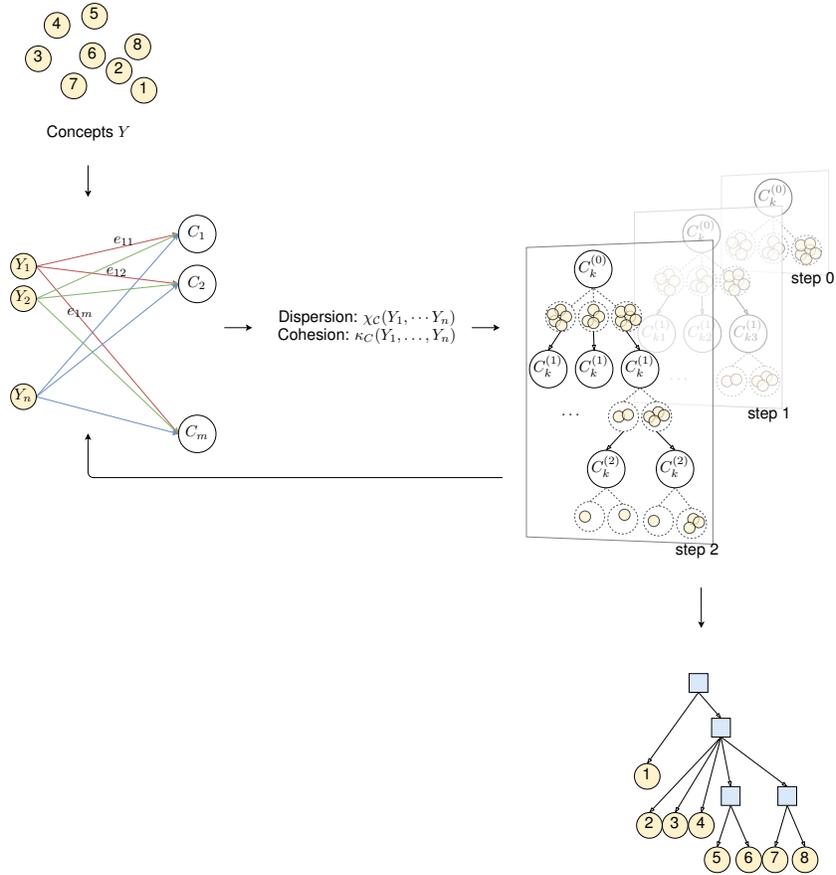
    }
}%
\caption{
    The framework of the proposed approach.
    Based on the dispersion and cohesion score obtained for each cluster, the best clustering solution is selected (step 0), and the process is repeated recursively on each group of concepts within the selected clustering solution (subsequent steps 1, 2, etc.).
    The process ends as soon as we get individual concepts on the leaves of the decomposition hierarchy.
    The final hierarchy guides the learning process, where the learner will be trained on the groups of concepts within their descendant leaves.
    Figure adapted from~\cite{hamidi2022metalearning}.
}
\label{fig:clustering-approach}
\end{figure}

This metalearning approach for structuring the learning process based on the concepts to learn was proposed in~\cite{osmani2022clustering}.
This original approach combines clustering and classification of groups of concepts based on two original measures. 
The guiding principle of this approach is that instances of closely related concepts naturally cluster together.
Precisely, two novel measures (dispersion and cohesion) are used to assess the quality of clustering solutions regarding concept separability. These measures are optimized throughout the process until an optimal learning hierarchy is derived.
Figure~\ref{fig:clustering-approach} illustrates the proposed approach.

\subsection{Concepts Structuring Based on Transfer Affinity}
\label{sec:concepts-structuring-based-on-transfer-affinity}
\begin{figure*}[h!]
\captionsetup[subfigure]{labelformat=empty}
\centering
\sffamily
\subfloat[]{
    \def\svgwidth{2.\columnwidth}
    \resizebox{110mm}{!}{
       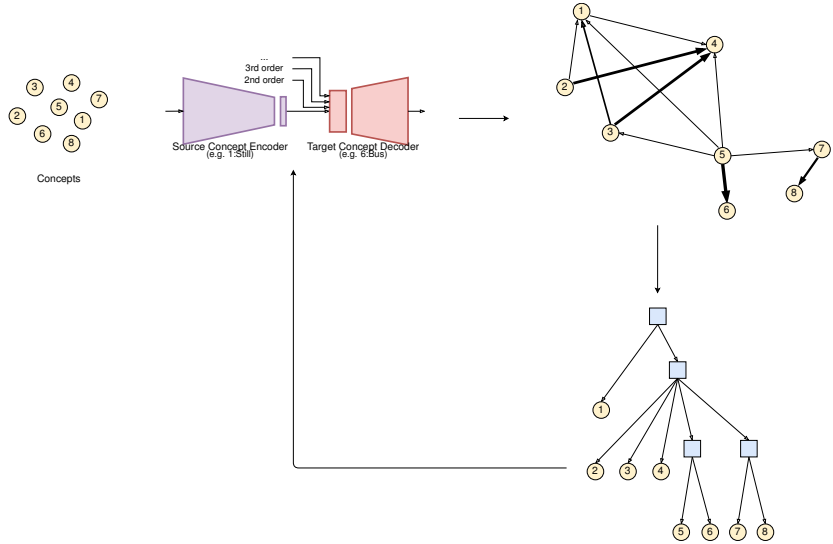
    }
}%
\caption{
    Our solution involves several repetitions of 3 main steps: (1) \textit{Concept similarity analysis}:
    encoders are trained to output an appropriate representation for each source concept, which is then fine-tuned to serve target concepts. Affinity scores are depicted by the arrows between concepts (the thicker the arrow, the higher the affinity score).
    (2) \textit{Hierarchy derivation}:
    based on the obtained affinity scores, a hierarchy is derived using the hierarchical agglomerate clustering approach.
    (3) \textit{Hierarchy refinement}: each non-leaf node of the derived hierarchy is assigned with a model that encompasses an appropriate representation as well as additional dense layers which are optimized to separate the considered concepts.
    Figure adapted from~\cite{hamidi2022metalearning}
}
\label{fig:proposed-hierarchical-approach}
\end{figure*}

The guiding principle here is to maximize transfer, sharing, and reuse while constructing the hierarchies.
The proposed approach in~\cite{osmani2021hierarchical} is still data-driven, but the considered concepts are structured in a bottom-up process instead of the top-down one presented above (§~\ref{sec:concepts-structuring-based-on-clustering}).
This approach is based on transfer affinity to determine an optimal organization of the concepts. This powerful technique based on transfer learning showed interesting empirical properties in various domains~\cite{zamir2018taskonomy,peters2019tune}.
This approach starts by computing concept dependencies that exist in the data domain using the transfer affinity scores. The closest concepts are then fused hierarchically with each other.
When taking a bottom-up process, the complete hierarchy, including the parameters assigned to each non-leaf node, can be learned incrementally by reusing what was learned on the way, i.e., while computing the transfer affinity scores.
See Figure~\ref{fig:proposed-hierarchical-approach} for an illustration of the proposed approach.

\subsection{Discussion}
\begin{figure}[h!]
    \centering
    \includegraphics[width=0.45\columnwidth]{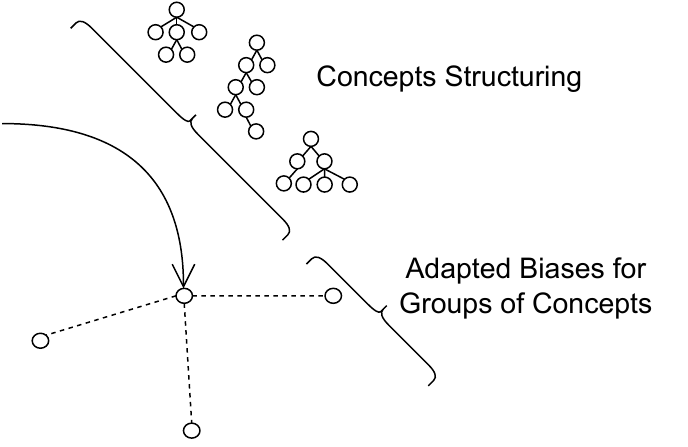}
    \caption{
        Optimizing for more adapted concepts structuring to tackle group biases.
    }
    \label{fig:MAML-vs-concept-hierarchies}
\end{figure}

\paragraph{Bias learning as either (i) the hierarchy construction process or (ii) the construction of a hierarchy of increasingly specialized biases}
The case of concept hierarchy is an instance of the meta-learning problem, and as such, it can be seen in several ways:
(i) as illustrated in Figure~\ref{fig:MAML-vs-concept-hierarchies}, where the upper level corresponds to the stages of construction of the most suitable (optimal) hierarchy, the lower level consisting of a process whose final goal is to adapt the weights of (or what is learned by) each of the nodes of the hierarchy;
(ii) the upper level does not correspond to the construction of an optimal hierarchy but is subdivided into several other levels that correspond to the different levels (groupings of concepts) of a given hierarchy (available a priori or built beforehand). In each of these levels, bias learning takes place on a group of concepts, which puts the following level in a good position to learn the bias of the level that follows it up to the level of atomic concepts.
In both cases, the higher level(s) correspond(s) to bias learning.






\section{Conclusion}
In this paper, we discussed metalearning, or the necessity of learning appropriate parameterizations for the learning processes.
This is even more important as the various biases that arise in real-world applications often lead to ill-conditioned learning problems that become utterly hard to solve.
To illustrate our discussion, we focused on the idea of guiding the learning processes based on how the concepts to learn are structured.
We presented two metalearning approaches that learn how to guide the learning processes.

\bibliography{biblio}

\begin{thebibliography}{31}
\providecommand{\natexlab}[1]{#1}
\providecommand{\url}[1]{\texttt{#1}}
\expandafter\ifx\csname urlstyle\endcsname\relax
  \providecommand{\doi}[1]{doi: #1}\else
  \providecommand{\doi}{doi: \begingroup \urlstyle{rm}\Url}\fi

\bibitem[Abnar et~al.(2020)Abnar, Dehghani, and Zuidema]{abnar2020transferring}
Samira Abnar, Mostafa Dehghani, and Willem Zuidema.
\newblock Transferring inductive biases through knowledge distillation.
\newblock 2020.

\bibitem[Ahmed et~al.(2019)Ahmed, Le~Roux, Norouzi, and
  Schuurmans]{ahmed2019understanding}
Zafarali Ahmed, Nicolas Le~Roux, Mohammad Norouzi, and Dale Schuurmans.
\newblock Understanding the impact of entropy on policy optimization.
\newblock In \emph{International conference on machine learning}, pp.\
  151--160. PMLR, 2019.

\bibitem[Banos et~al.(2014)Banos, Galvez, Damas, Pomares, and
  Rojas]{banos2014window}
Oresti Banos, Juan-Manuel Galvez, Miguel Damas, Hector Pomares, and Ignacio
  Rojas.
\newblock Window size impact in human activity recognition.
\newblock \emph{Sensors}, 14\penalty0 (4):\penalty0 6474--6499, 2014.

\bibitem[Battaglia et~al.(2018)Battaglia, Hamrick, Bapst, Sanchez-Gonzalez,
  Zambaldi, Malinowski, Tacchetti, Raposo, Santoro, Faulkner,
  et~al.]{battaglia2018relational}
Peter~W Battaglia, Jessica~B Hamrick, Victor Bapst, Alvaro Sanchez-Gonzalez,
  Vinicius Zambaldi, Mateusz Malinowski, Andrea Tacchetti, David Raposo, Adam
  Santoro, Ryan Faulkner, et~al.
\newblock Relational inductive biases, deep learning, and graph networks.
\newblock \emph{arXiv preprint arXiv:1806.01261}, 2018.

\bibitem[Cully et~al.(2015)Cully, Clune, Tarapore, and Mouret]{cully2015robots}
Antoine Cully, Jeff Clune, Danesh Tarapore, and Jean-Baptiste Mouret.
\newblock Robots that can adapt like animals.
\newblock \emph{Nature}, 521\penalty0 (7553):\penalty0 503--507, 2015.

\bibitem[Dauphin et~al.(2014)Dauphin, Pascanu, Gulcehre, Cho, Ganguli, and
  Bengio]{dauphin2014identifying}
Yann~N Dauphin, Razvan Pascanu, Caglar Gulcehre, Kyunghyun Cho, Surya Ganguli,
  and Yoshua Bengio.
\newblock Identifying and attacking the saddle point problem in
  high-dimensional non-convex optimization.
\newblock \emph{Advances in neural information processing systems}, 27, 2014.

\bibitem[Finn et~al.(2017)Finn, Abbeel, and Levine]{finn2017model}
Chelsea Finn, Pieter Abbeel, and Sergey Levine.
\newblock Model-agnostic meta-learning for fast adaptation of deep networks.
\newblock In \emph{ICML}, 2017.

\bibitem[Gaier \& Ha(2019)Gaier and Ha]{gaier2019weight}
Adam Gaier and David Ha.
\newblock Weight agnostic neural networks.
\newblock In \emph{Advances in Neural Information Processing Systems}, pp.\
  5364--5378, 2019.

\bibitem[Hamidi(2022)]{hamidi2022metalearning}
Massinissa Hamidi.
\newblock \emph{Metalearning guided by domain knowledge in distributed and
  decentralized applications}.
\newblock PhD thesis, Universit{\'e} Paris-Nord-Paris XIII, 2022.

\bibitem[Hamidi \& Osmani(2020)Hamidi and Osmani]{hamidi2020data}
Massinissa Hamidi and Aomar Osmani.
\newblock Data generation process modeling for activity recognition.
\newblock In \emph{European Conference on Machine Learning and Principles and
  Practice of Knowledge Discovery in Databases}. Springer, 2020.

\bibitem[Hamidi \& Osmani(2021)Hamidi and Osmani]{hamidi2021human}
Massinissa Hamidi and Aomar Osmani.
\newblock Human activity recognition: A dynamic inductive bias selection
  perspective.
\newblock \emph{Sensors}, 21\penalty0 (21):\penalty0 7278, 2021.

\bibitem[Hamidi \& Osmani(2022)Hamidi and Osmani]{hamidi2022context}
Massinissa Hamidi and Aomar Osmani.
\newblock Context abstraction to improve decentralized machine learning in
  structured sensing environments.
\newblock In \emph{European Conference on Machine Learning and Principles and
  Practice of Knowledge Discovery in Databases}. Springer, 2022.

\bibitem[Hamidi et~al.(2020)Hamidi, Osmani, and Alizadeh]{hamidi2020multi}
Massinissa Hamidi, Aomar Osmani, and Pegah Alizadeh.
\newblock A multi-view architecture for the shl challenge.
\newblock UbiComp-ISWC '20, pp.\  317–322, New York, NY, USA, 2020.
  Association for Computing Machinery.

\bibitem[Hammerla \& Pl{\"o}tz(2015)Hammerla and Pl{\"o}tz]{hammerla2015let}
Nils~Y Hammerla and Thomas Pl{\"o}tz.
\newblock Let's (not) stick together: pairwise similarity biases
  cross-validation in activity recognition.
\newblock In \emph{Proceedings of the 2015 ACM international joint conference
  on pervasive and ubiquitous computing}, pp.\  1041--1051, 2015.

\bibitem[Kone{\v{c}}n{\`y} et~al.(2016)Kone{\v{c}}n{\`y}, McMahan, Ramage, and
  Richt{\'a}rik]{konevcny2016federated}
Jakub Kone{\v{c}}n{\`y}, H~Brendan McMahan, Daniel Ramage, and Peter
  Richt{\'a}rik.
\newblock Federated optimization: Distributed machine learning for on-device
  intelligence.
\newblock \emph{arXiv preprint arXiv:1610.02527}, 2016.

\bibitem[Li et~al.(2018)Li, Xu, Taylor, Studer, and
  Goldstein]{li2018visualizing}
Hao Li, Zheng Xu, Gavin Taylor, Christoph Studer, and Tom Goldstein.
\newblock Visualizing the loss landscape of neural nets.
\newblock \emph{Advances in neural information processing systems}, 31, 2018.

\bibitem[Li et~al.(2020)Li, Sahu, Zaheer, Sanjabi, Talwalkar, and
  Smith]{li2020federated}
Tian Li, Anit~Kumar Sahu, Manzil Zaheer, Maziar Sanjabi, Ameet Talwalkar, and
  Virginia Smith.
\newblock Federated optimization in heterogeneous networks.
\newblock \emph{MLSys}, 2:\penalty0 429--450, 2020.

\bibitem[Lin et~al.(2021)Lin, Wang, Sun, Chen, Sun, Qian, Li, and
  Jin]{lin2021zen}
Ming Lin, Pichao Wang, Zhenhong Sun, Hesen Chen, Xiuyu Sun, Qi~Qian, Hao Li,
  and Rong Jin.
\newblock Zen-nas: A zero-shot nas for high-performance deep image recognition.
\newblock \emph{arXiv preprint arXiv:2102.01063}, 2021.

\bibitem[Mirzadeh et~al.(2020)Mirzadeh, Farajtabar, Gorur, Pascanu, and
  Ghasemzadeh]{mirzadeh2020linear}
Seyed~Iman Mirzadeh, Mehrdad Farajtabar, Dilan Gorur, Razvan Pascanu, and
  Hassan Ghasemzadeh.
\newblock Linear mode connectivity in multitask and continual learning.
\newblock In \emph{International Conference on Learning Representations}, 2020.

\bibitem[Nichol \& Schulman(2018)Nichol and Schulman]{nichol2018reptile}
Alex Nichol and John Schulman.
\newblock Reptile: a scalable metalearning algorithm.
\newblock \emph{arXiv preprint arXiv:1803.02999}, 2\penalty0 (3):\penalty0 4,
  2018.

\bibitem[Osmani \& Hamidi(2022)Osmani and Hamidi]{osmani2022reduction}
Aomar Osmani and Massinissa Hamidi.
\newblock Reduction of the position bias via multi-level learning for activity
  recognition.
\newblock In \emph{Pacific-Asia Conference on Knowledge Discovery and Data
  Mining}, pp.\  289--302. Springer, 2022.

\bibitem[Osmani et~al.(2017{\natexlab{a}})Osmani, Hamidi, and
  Chibani]{osmani2017machine}
Aomar Osmani, Massinissa Hamidi, and Abdelghani Chibani.
\newblock Machine learning approach for infant cry interpretation.
\newblock In \emph{Tools with Artificial Intelligence (ICTAI), 2017 IEEE 29th
  International Conference on}, pp.\  182--186. IEEE, 2017{\natexlab{a}}.

\bibitem[Osmani et~al.(2017{\natexlab{b}})Osmani, Hamidi, and
  Chibani]{osmani2017platform}
Aomar Osmani, Massinissa Hamidi, and Abdelghani Chibani.
\newblock Platform for assessment and monitoring of infant comfort.
\newblock In \emph{2017 {AAAI} Fall Symposia, Arlington, Virginia, USA,
  November 9-11, 2017}, pp.\  36--44, 2017{\natexlab{b}}.

\bibitem[Osmani et~al.(2021)Osmani, Hamidi, and
  Alizadeh]{osmani2021hierarchical}
Aomar Osmani, Massinissa Hamidi, and Pegah Alizadeh.
\newblock Hierarchical learning of dependent concepts for human activity
  recognition.
\newblock In \emph{Pacific-Asia Conference on Knowledge Discovery and Data
  Mining}, pp.\  79--92. Springer, 2021.

\bibitem[Osmani et~al.(2022)Osmani, Hamidi, and Alizadeh]{osmani2022clustering}
Aomar Osmani, Massinissa Hamidi, and Pegah Alizadeh.
\newblock Clustering approach to solve hierarchical classification problem
  complexity.
\newblock In \emph{Proceedings of the AAAI Conference on Artificial
  Intelligence}, volume~36, pp.\  7904--7912, 2022.

\bibitem[Peters et~al.(2019)Peters, Ruder, and Smith]{peters2019tune}
Matthew~E Peters, Sebastian Ruder, and Noah~A Smith.
\newblock To tune or not to tune? adapting pretrained representations to
  diverse tasks.
\newblock \emph{arXiv preprint arXiv:1903.05987}, 2019.

\bibitem[Poggio \& Girosi(1990)Poggio and Girosi]{poggio1990regularization}
Tomaso Poggio and Federico Girosi.
\newblock Regularization algorithms for learning that are equivalent to
  multilayer networks.
\newblock \emph{Science}, 247\penalty0 (4945):\penalty0 978--982, 1990.

\bibitem[Raghu et~al.(2019)Raghu, Raghu, Bengio, and Vinyals]{raghu2019rapid}
Aniruddh Raghu, Maithra Raghu, Samy Bengio, and Oriol Vinyals.
\newblock Rapid learning or feature reuse? towards understanding the
  effectiveness of maml.
\newblock In \emph{International Conference on Learning Representations}, 2019.

\bibitem[Shoaib et~al.(2016)Shoaib, Bosch, Incel, Scholten, and
  Havinga]{shoaib2016complex}
Muhammad Shoaib, Stephan Bosch, Ozlem~Durmaz Incel, Hans Scholten, and Paul~JM
  Havinga.
\newblock Complex human activity recognition using smartphone and wrist-worn
  motion sensors.
\newblock \emph{Sensors}, 16\penalty0 (4):\penalty0 426, 2016.

\bibitem[Zamir et~al.(2018)Zamir, Sax, Shen, Guibas, Malik, and
  Savarese]{zamir2018taskonomy}
Amir~R Zamir, Alexander Sax, William Shen, Leonidas~J Guibas, Jitendra Malik,
  and Silvio Savarese.
\newblock Taskonomy: Disentangling task transfer learning.
\newblock In \emph{Proceedings of the IEEE conference on computer vision and
  pattern recognition}, pp.\  3712--3722, 2018.

\bibitem[Zhang et~al.(2020)Zhang, Qu, and Wright]{zhang2020symmetry}
Yuqian Zhang, Qing Qu, and John Wright.
\newblock From symmetry to geometry: Tractable nonconvex problems.
\newblock \emph{arXiv preprint arXiv:2007.06753}, 2020.

\end{thebibliography}
\bibliographystyle{iclr2024_conference}

\end{document}